\documentclass[10pt,twocolumn,letterpaper]{article}

\usepackage{cvpr}
\usepackage{times}
\usepackage{epsfig}
\usepackage{graphicx}
\usepackage{amsmath}
\usepackage{amssymb}
\usepackage{subfig}

\usepackage[pagebackref=true,breaklinks=true,letterpaper=true,colorlinks,bookmarks=false]{hyperref}

\cvprfinalcopy 

\def\cvprPaperID{****} 

\ifcvprfinal\fi

\newtheorem{thm}{Theorem}

\begin{document}

\title{A Modular and Unified Framework for Detecting and Localizing Video Anomalies}

\author{Keval Doshi\\
University of South Florida\\
4202 E Fowler Ave, Tampa, FL 33620\\
{\tt\small kevaldoshi@usf.edu}
\and
Yasin Yilmaz\\
University of South Florida\\
4202 E Fowler Ave, Tampa, FL 33620\\
{\tt\small yasiny@usf.edu}
}

\maketitle

\begin{abstract}
Anomaly detection in videos has been attracting an increasing amount of attention. Despite the competitive performance of recent methods on benchmark datasets, they typically lack desirable features such as modularity, cross-domain adaptivity, interpretability, and real-time anomalous event detection. Furthermore, current state-of-the-art approaches are evaluated using the standard instance-based detection metric by considering video frames as independent instances, which is not ideal for video anomaly detection. 
Motivated by these research gaps, we propose a modular and unified approach to the online video anomaly detection and localization problem, called MOVAD, which consists of a novel transfer learning based plug-and-play architecture, a sequential anomaly detector, a mathematical framework for selecting the detection threshold, and a suitable performance metric for real-time anomalous event detection in videos. Extensive performance evaluations on benchmark datasets show that the proposed framework significantly outperforms the current state-of-the-art approaches.
\end{abstract}
\vspace{-2mm}
\vspace{-2mm}
\section{Intoduction}

With the increasing demand for security, increasing storage and processing capabilities, and decreasing cost of electronics, surveillance
cameras have been widely deployed 
\cite{zhou2019anomalynet}. Due to the exponential increase in the number of CCTV cameras, the amount of video generated far surpasses our ability to manually analyze it. Automated detection of anomalies in video is challenging since the definition of “anomaly” is ambiguous -- any event that does not conform to “normal” behaviors can be considered as an anomaly. For example, a person riding a bike is usually a nominal behavior, however, it may be considered as anomalous if it occurs in a restricted space.

Specifically, due to the important role video anomaly detection plays in ensuring safety, security, and sometimes prevention of potential catastrophes, a major functionality of a video anomaly detection system is the real-time decision making capability. While there is a lot of prior work on anomaly detection in surveillance videos, they mainly \emph{focus on offline localization of anomaly in video frames} following an instance-based binary hypothesis testing approach and \emph{ignoring the online (i.e., real-time) detection of anomalous events}. For example, most of the existing works, e.g. \cite{ionescu2019object,liu2018future,zhou2019anomalynet}, employ a video normalization technique that requires an entire video segment for computation. They also typically depend on the assumption that there is an anomaly in the video segment. In practice, this assumption either will not hold for short video segments (on the order of minutes) or will cause long delays in detecting anomalous events for sufficiently long video segments (on the order of days). 

The automated video surveillance literature lacks a clear distinction between online anomalous event detection and offline anomalous frame localization \cite{liu2018future,ionescu2019object,ramachandra2020street,pang2020self,markovitz2020graph}. While the commonly used frame-level AUC (area under the ROC curve), which is borrowed from the instance-based binary hypothesis testing, might be a suitable metric for localizing the anomaly in video frames, it ignores the temporal nature of videos and fails to capture the dynamics of detection results, e.g., a detector that detects a late portion of an anomalous event and alarms the user after a long delay can achieve the same frame-level AUC as the detector that quickly detects the anomalous event and timely alarms the user but misses some anomalous frames afterwards. 
While minimizing the delay in detecting an anomalous event is critical \cite{mao2019delay}, it is also necessary to control the false alarm rate. Hence, a video anomaly detector should aim to judiciously raise alarms in a timely manner.

For practical implementations, it is unrealistic to assume the availability of sufficient training data such that it encompasses all possible nominal events/behaviors. Thus, a practical framework should also be able to perform \emph{few-shot} adaptation to new nominal scenarios over time. This presents a novel challenge to the current approaches discussed in Section \ref{s:related} as their decision functions heavily depend on Deep Neural Networks (DNNs) \cite{doshi2020continual}. DNNs typically require a large amount of training data to learn a new nominal pattern or exhibit the risk of catastrophic forgetting with incremental updates \cite{kirkpatrick2017overcoming}. 

Another limitation of existing methods is the lack of interpretability due to the inclination towards end-to-end deep learning based models, leading to a semantic gap between the visual features and the real interpretation of events \cite{morais2019learning}. While such models perform well on some benchmark datasets, i.e., they are easily able to detect a certain category of anomalies, they cannot adequately generalize to other types of anomalies. For example, \cite{morais2019learning,rodrigues2020multi,markovitz2020graph} propose a pose estimation based framework, and hence are only able to detect human-related anomalies. Moreover, there is no straightforward way to modify such methods to target a different class of anomaly since they are based on intricately designed neural networks.  

Our goal in this paper is to present a more systematic framework for video anomaly detection and localization, and tackle practical challenges such as few-shot adaptation, which is largely unexplored in the existing literature. In summary, our contributions in this paper are as follows:
\begin{itemize}
    \item We present a systematic unified framework for online event detection and offline frame localization for video anomalies, and propose a new performance metric for online event detection. 
    \item We propose a modular transfer learning based anomaly detection architecture which can be easily modified to target specific anomaly categories and can easily adapt to new scenarios using a few samples (cross-domain adaptivity). 
    \item We introduce a statistical technique for the selection of detection threshold to satisfy a desired false alarm rate. 
\end{itemize}


\section{Related Works}
\label{s:related}

There is a fast-growing body of research investigating anomaly detection in videos. A key component of computer vision problems is the extraction of meaningful features. In video surveillance, the extracted features should be capable of capturing the difference between nominal and anomalous events within a video \cite{doshi2020continual}. While some methods use supervised learning to train on both nominal and anomalous events \cite{liu2019exploring,landi2019anomaly}, the majority of existing research is concentrated on semi-supervised learning due to the limitations in the availability of annotated anomalous instances. Early anomaly detection methods used handcrafted approaches which extract different types of motion information in the form of histogram of oriented gradients (HOGs) \cite{chaudhry2009histograms,colque2016histograms} and optical flow. Another category is sparse coding-based methods \cite{zhao2011online}, which were used to learn a dictionary of normal sparse events, and attempt to detect anomalies based on the reconstructability of video from the dictionary atoms. For example, \cite{mo2013adaptive} uses sparse reconstruction to learn joint trajectory representations of multiple objects. These approaches, while computationally inexpensive, often fail to capture complex anomalous patterns. The recent literature however has been dominated by Convolutional Neural Network (CNN) based methods \cite{hasan2016learning,hinami2017joint,luo2017revisit,ravanbakhsh2019training,sabokrou2018adversarially,xu2015learning,markovitz2020graph,ramachandra2020street,ionescu2019object} due to their significantly superior detection performance. Recently, transfer learning based object detection methods have also been frequently used \cite{doshi2020any,doshi2020continual,ionescu2019object,georgescu2020scene} to learn appearance features. The neural network-based methods can be broadly segregated into reconstruction-based methods \cite{hasan2016learning,ravanbakhsh2017abnormal,chong2017abnormal,ionescu2019object} and prediction-based methods \cite{liu2018future,lu2020few,ruff2018deep}. However, these CNNs require a significant amount of training to adapt to a new scenario. Hence, recently few-shot learning has been gaining attention in the computer vision literature \cite{koch2015siamese,sung2018learning,snell2017prototypical,vinyals2016matching,lomonaco2017core50,lu2020few}. However, no significant progress has been made yet in few-shot scene adaptation for video surveillance. Hence, in this work, we primarily compare our few-shot adaptation performance with \cite{lu2020few}, which proposes a meta-learning algorithm for cross-domain adaptivity.  

\section{Proposed Method}
\label{s:proposed}

\begin{figure*}[!th]
\centering
\includegraphics[width=1\textwidth]{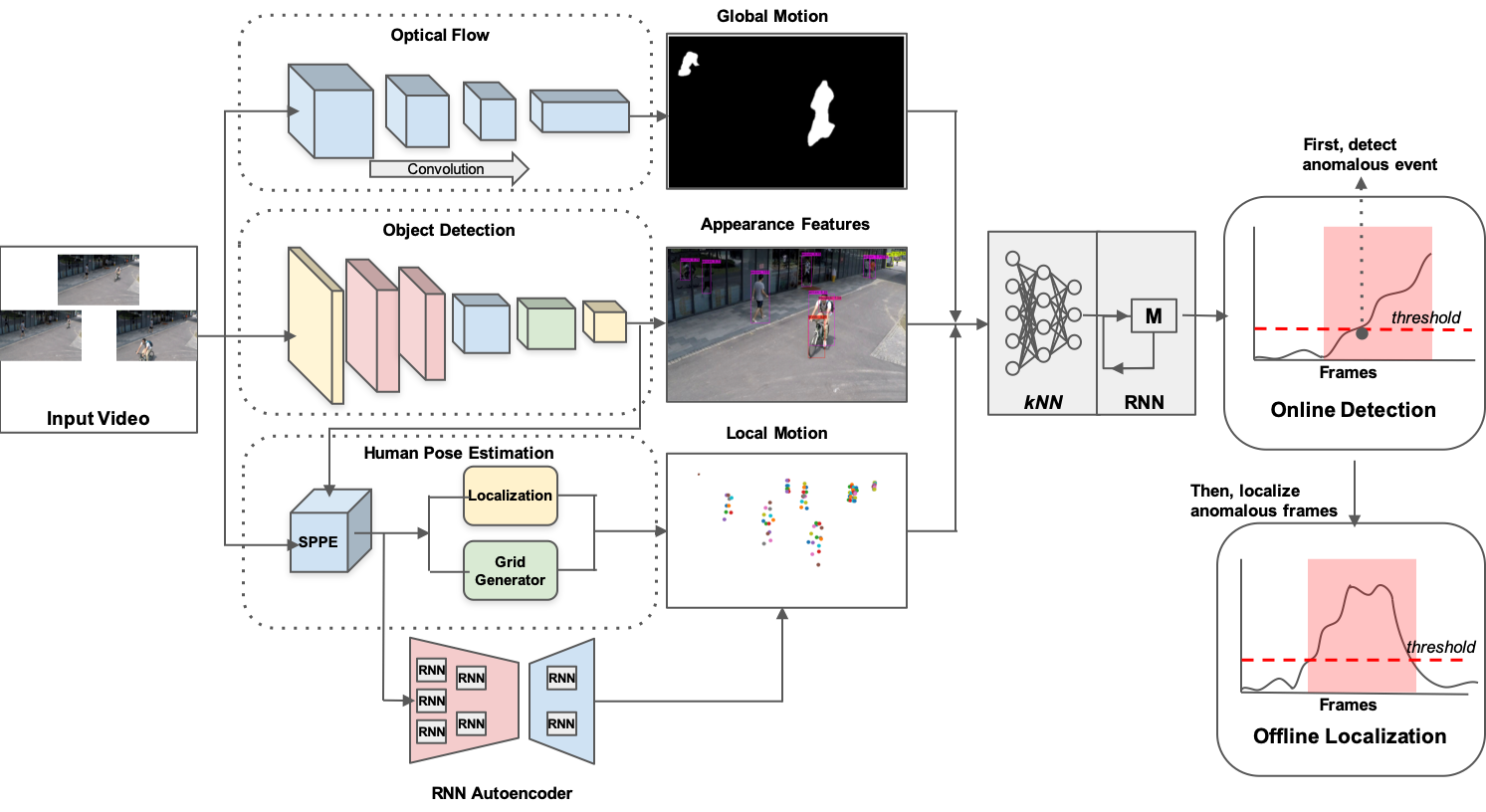}
\vspace{-2mm}
\caption{Proposed MOVAD framework. At each time $t$, neural network-based feature extraction module provides location (center coordinates and area of bounding box), appearance (class probabilities), global motion (optical flow), and local motion (pose estimation) features to the statistical anomaly detection module, which computes $k$NN distance for anomaly evidence using a fully connected neural network, and sequentially decides for anomalous events using an RNN. In human pose estimation, the single person pose estimation (SPPE) is converted to multi-person pose features.}
\label{f:system}
\vspace{-2mm}
\end{figure*}

\subsection{Motivation}

In the recent anomaly detection literature, most of the proposed methods consist of training a deep neural network on available nominal samples. However, such an approach has several shortcomings. First, the applicability of such a method is limited to a few scenarios where there is a drastic change in the appearance or motion of an object. In \cite{doshi2020any}, it is shown that modifying the benchmark datasets results in a significant drop in the performance of state-of-the-art algorithms. Second, to the best of our knowledge, there is no existing method that can be easily modified or extended to a new category of anomalies. For example, even recent algorithms such as \cite{zaheer2020old,liu2018future,park2020learning} cannot detect (or be modified to detect) anomalies pertaining to changes in human poses. Third, because of the extensive use of end-to-end learning in recent algorithms, the models lack interpretability. While there are certain supervised methods, e.g., \cite{sultani2018real}, which are capable of recognizing the type of anomaly, they depend on the availability of anomalous data. Finally, existing methods also lack a clear procedure for incorporating new knowledge, and would likely necessitate significant changes to the existing architecture. 

Motivated by these shortcomings, we propose a modular framework, called \emph{Modular Online Video Anomaly Detector (MOVAD)}, consisting of deep learning-based feature extraction and statistical anomaly detection, as shown in Fig. \ref{f:system}. In particular, transfer learning based convolutional neural networks (CNNs) and recurrent neural networks (RNNs) are used to extract informative features, followed by a novel $k$NN-based neural network and RNN-based sequential anomaly detector. 

The choice of separating feature extraction module and decision module also enables theoretical performance analysis and a closed-form expression for the detection threshold. 
In the following sections, we discuss our framework in detail.    

\subsection{Transfer Learning-Based Feature Extraction}
In general, the end-to-end training of DNNs for video anomaly detection necessitates focusing on a particular aspect in which anomalies may occur, such as object appearance or motion or pose, and extracting only those features. However, even in the same scene, anomalous events may be manifested in different aspects. Hence, advanced video anomaly detectors should utilize features from multiple aspects together. For instance, biological vision systems extracts different features in the visual cortex such as appearance, global motion, and local motion \cite{biological}. To this end, we propose a flexible feature extraction module that can work with various modalities, which enables a plug-and-play modular architecture. This means although appearance, global motion, and local motion features are considered in this paper, the proposed framework can be easily modified to add new feature extractors or remove existing ones. 
Furthermore, entirely retraining a video anomaly detector for new scene/domain is typically not necessary since most domains share the same feature types (appearance, global motion, local motion, etc.). As a result, to significantly reduce the training computational complexity, a transfer learning approach is utilized in the proposed framework. We next explain the considered feature extractors, which work in parallel as shown in Fig. \ref{f:system}. 

\label{s:detect}

\textbf{Object Appearance:} 
A pre-trained object detection system is used to detect objects and extract appearance and spatial features. Since we do not assume any prior knowledge about the type of anomalies, and hence by extension the object classes, we use a model trained on the MS-COCO dataset. For online anomaly detection, the real-time operation is a critical factor, and hence, we currently prefer the You Only Look Once (YOLO) \cite{redmon2016you} algorithm, specifically YOLOv4, in our implementations. It should be noted that the choice of the object detector is not critical for the proposed framework, and can be adjusted according to the application. Using the object detector, we extract the bounding box (location) as well as the class probabilities (appearance) for each object detected in a given frame. Instead of directly using the bounding box coordinates, we instead compute the center and area of the box and leverage them as our spatial features. During testing, any object belonging to a previously unseen class and/or deviating from the known nominal paths contributes to an anomalous event alarm.             

\textbf{Global Motion:} Apart from spatial and appearance features, capturing the motion of different objects is also critical for detecting anomalies in videos. Hence, to monitor the contextual motion of different objects, we propose using a pre-trained optical flow model such as Flownet 2 \cite{ilg2017flownet}. We hypothesize that objects with an unusually high/low optical flow intensity would exhibit an anomalous behavior. 
Thus, the mean and variance are for each detected object are used as our global motion features.    

\textbf{Local Motion:} To study the social behavior in a video, it is an important factor to study the human motion closely. For inanimate objects like cars, trucks, bikes, etc., monitoring the optical flow is sufficient to judge whether they portray some sort of anomalous behavior. However, with regard to humans, we also need to monitor their poses to determine whether an action is anomalous or not. Hence, using a pre-trained multi-person pose estimator such as AlphaPose \cite{fang2017rmpe} is proposed to extract skeletal trajectories.

\subsection{Statistical Anomaly Detection}
\label{s:sad}

\textbf{Anomaly Evidence:} 
Given the various extracted features, the next step in the proposed framework is to compute an anomaly evidence score for each video frame in an online fashion. Due its favorable characteristics, such as interpretability and theoretical tractability, we use $k$-nearest-neighbor ($k$NN) distance as an anomaly evidence. 
For a feature vector $X_{t,i} \in \mathbb{R}^m$ representing each object $i$ in frame $t$, our objective is to compute its Euclidean distance $D_{t,i}$ to the $k$th nearest feature vector in the nominal training set. Since $k$NN distance computation becomes expensive with increasing training size, for scalability, we propose training a fully connected neural network with parameters $\theta$, which takes $X_{t,i}$ as the input and gives an accurate approximation $\tilde{D}_{t,i}(\theta)$ to $D_{t,i}$. The objective function for training the $k$NN neural network is given by
\begin{equation}
    \min_\theta \frac{1}{N}\sum_{j=1}^N (D_j-\tilde{D}_j(\theta))^2 + \lambda f(\theta),
\end{equation}
where $N$ is the number of feature vectors in the training set, $\lambda f(\theta)$ is the regularization term. The number of neighbors $k$ determines a trade-off between sensitivity to anomalies and robustness to nominal outliers. While smaller $k$ values makes the system more sensitive to real anomalies, it may also make the system more vulnerable to nominal outliers. However, the choice of $k$ is not critical for the detection performance since the proposed sequential detection module does not directly decide on the anomaly evidences. As shown next, through the internal memory of the RNN structure, it gathers the evidences to detect anomalous events, hence does not typically raise an alarm due to a single evidence due to an outlying frame. 

\textbf{Online Anomaly Detection:}
To accommodate the temporal continuity of video data and detect anomalous events in an online fashion, a sequential statistical decision making method based on RNN is proposed. The anomaly evidence scores (i.e., $k$NN distances) from streaming video frames provide an informative time series data which typically takes large values when the anomalous event starts. However, to avoid false alarms due to outlying large evidences from nominal frames, the proposed framework does not decide using individual evidences, but instead utilizes the temporal information inherent in the evidence time series (i.e., an anomalous event consists of a number of successive anomalous frames). Specifically, it takes the streaming $k$NN distances $\{\tilde{D}_t\}$ as input and updates an internal state, which is then passed through ReLU activation function to yield the decision statistic $s_t$. The time series $\{\tilde{D}_t\}$ is obtained by taking the largest $k$NN distance among objects in each frame, i.e., $\tilde{D}_t=\max_i \tilde{D}_{t,i}$. The output neuron in RNN compares $s_t$ with a threshold $h$ to raise an alarm if $s_t\ge h$ or continue with the next frame otherwise. Note that the RNN structure can be expanded to accept multiple time series (in addition to $k$NN distances) and to have deeper layers if desired. While $k$NN distances are available for the nominal class, there is no such scores for the anomaly class to train RNN in the considered semi-supervised setup. Synthetic $k$NN distances are generated uniformly in the interval $(D_{\alpha},2D_{max})$ where $\alpha$ is a statistical significance level (e.g., $\alpha=0.05$), $D_{\alpha}$ is the $(1-\alpha)$ percentile of nominal distances in the training set, and $D_{max}$ is the maximum nominal distance in the training. 

To circumvent the training with synthetic data, and obtain a closed-form expression for the threshold $h$, we also propose a simplified decision rule. Motivated by the resemblance of the memory (internal state) and ReLU operations of RNN with the minimax optimum sequential change detection algorithm CUSUM \cite{basseville1993detection}, we consider fixing the RNN weights to obtain the simplified decision statistic $\tilde{s}_t=\max\{\tilde{s}_{t-1}+\delta_t,0\}$. In this update rule, the weights of internal state and input are set to one, where the input 
$\delta_t=\tilde{D}_t^m-D_{\alpha}^m$ is the normalized $k$NN distance, where $m$ is the dimensionality of feature vectors $X_{t,i}$. 
In our experiments, the simplified detector gave very similar results to the general RNN detector. With the weights set to one, there is no need to train the RNN, and the simplified decision statistic $\tilde{s}_t$ lends itself to theoretical analysis to derive a closed-form expression for the threshold $h$, as explained next. 

\begin{thm}
\label{thm:thr}
As the training size grows ($N\to\infty$), the false alarm rate of the proposed simplified detector based on $\tilde{s}_t$ is upper bounded by $FAR \le e^{-\omega_0 h}$ and the threshold $h$ can be set as 
\begin{equation}
\label{e:thr}
h=\frac{-\log \beta}{\omega_0}
\end{equation}
to asymptotically satisfy a desired false alrm constraint $FAR \le \beta$. The constant $\omega_0$ is computed from the training data and given by
\begin{align}
    \label{e:thm}
    \omega_0 &= v_m - \theta -\frac{1}{\phi} \mathcal{W}\left( -\phi \theta e^{-\phi\theta } \right), \\
    \theta &= \frac{v_m}{e^{v_m D_\alpha^m}},\nonumber
\end{align}
where $\mathcal{W}(\cdot)$ is the Lambert-W function, $v_m=\frac{\pi^{m/2}}{\Gamma(m/2+1)}$ is the constant for the $m$-dimensional Lebesgue measure (i.e., $v_m d_\alpha^m$ is the $m$-dimensional volume of the hyperball with radius $d_\alpha$), and $\phi$ is the upper bound for $\delta_t$.
\end{thm}
\textit{Proof.} See the supplementary file.

Although the expression for $\omega_0$ looks complicated, all the terms in Eq. \eqref{e:thm} can be easily computed. Particularly, $v_m$ is directly given by the number of features $m$, $D_\alpha$ comes from the training phase, $\phi$ is also found in training, and finally there is a built-in Lambert-W function in popular programming languages such as Python and Matlab. 
Hence, given the training data, $\omega_0$ can be easily computed, and the threshold $h$ can be chosen using Eq. \eqref{e:thr} to asymptotically achieve the desired false alarm rate $\beta\in(0,1)$. 

Decision threshold $h$ is a key parameter that is common to all existing anomaly detection algorithms, and yet is often overlooked. Since an alarm is raised when the test statistic crosses the threshold, choosing an appropriate threshold is critical for controlling the number of false alarms and minimizing the need for human involvement. In a practical setting, without a clear procedure for selecting the decision threshold, an exhaustive empirical process is needed to calibrate the threshold for an acceptable false alarm rate.

\textbf{New Performance Metric for Online Detection:} Low detection delay is a crucial requirement in most video-related applications such as autonomous driving \cite{lin2018architectural} and automated video surveillance. However, the detection delay, which is the time required by an algorithm to detect an anomalous event, is largely unexplored in the field of video anomaly detection. 
The popular performance metric in the video anomaly detection literature, AUC, cannot effectively evaluate the performance of online anomaly detection algorithms \cite{lavin2015evaluating}. Hence, we present a new performance metric called APD (Average Precision as a function of Delay), which is based on average detection delay and precision. The proposed delay metric is given by 

\begin{equation}
\label{eq:met}
    \text{APD} = \int_{0}^{1} P(\gamma) ~\text{d}\gamma,
\end{equation}
where $\gamma$ denotes the normalized average detection delay, and $P$ denotes the precision. The average detection delay is normalized by the largest possible delay either defined by a performance requirement or the length of natural cuts in the video stream such as the video segments in the benchmark datasets (See Sec. \ref{s:dataset}).

\textbf{Offline Localization:} 
Once an anomalous event is detected, the detection instance is marked as the starting point, and the decision statistic is updated as usual to determine the end point. When the decision statistic drops consecutively for a number of frames (e.g., five frames is found to be a good number in our experiments), the beginning of the drop window is marked as the end point. Finally, the frames between the start and end points are labeled as anomalous.

\textbf{Implementation Details:} In our implementation, we fix the number of neighbors as $k=10$. However, as indicated in Section \ref{s:sad}, the choice of $k$ is not sensitive and does not significantly affect the performance of the detector. The detection performance is controlled by the decision threshold $h$, which can be mathematically set by following Eq. \eqref{e:thr}. For the $k$NN regression network, we use a fully connected deep neural network with 3 hidden layers consisting of 20 neurons each. We empirically chose the simplest network that gave a sufficiently low prediction error. The feature vector is 18-dimensional for each detected object, and consists of 15 class probabilities (appearance), mean and variance of optical flow in the bounding box (global motion), and prediction error of pose if human (local motion). Global and local motion features are normalized to [0,1] using the min and max values from the training data. 

\section{Experiments}

\label{s:experiment}

In this section, we first briefly discuss the benchmark datasets and the evaluation metrics. Then, we provide a detailed comparison between the proposed algorithm and the state-of-the-art algorithms in terms of online detection and offline localization. We also evaluate our few-shot adaptation performance.

\subsection{Datasets}
\label{s:dataset}

We consider four publicly available benchmark datasets, namely the CUHK Avenue dataset, the UCSD pedestrian dataset, the ShanghaiTech campus dataset, and the UR fall dataset.  

\textbf{ UCSD Ped 2}: The UCSD pedestrian dataset is one of the most widely used video anomaly detection datasets. Due to the low resolution of the UCSD Ped 1 videos, we only consider the UCSD Ped 2 dataset. The Ped 2 dataset consists of 16 training videos and 12 test videos. The anomalous events are caused due to vehicles such as bicycles, skateboards and wheelchairs. Despite being widely used as a benchmark dataset, most anomalies are obvious and can be easily detected from a single frame.  

\textbf{CUHK Avenue:} Another popular dataset is the CUHK Avenue dataset, which consists of short video clips taken from a single outdoor surveillance camera looking at the side of a building with a pedestrian walkway in front of it. It contains 16 training and 21 test videos with a frame resolution of 360 $\times$ 640.

\textbf{ShanghaiTech:} The ShanghaiTech dataset is one of the largest and most challenging datasets available for anomaly detection in videos.  It consists of 330 training and 107 test videos from 13 different scenes, which sets it apart from the other available datasets. The resolution for each video frame is 480 $\times$ 856.

\textbf{UR Fall}: While the UR fall dataset is not popularly used for video anomaly detection, it has recently been proposed for testing the generalization capability of anomaly detection algorithms \cite{lu2020few}. This dataset contains 70 depth videos collected with a Microsoft Kinect camera in a nursing home and the anomalies consist of a person falling in a closed room.

\subsection{Results}

\begin{figure}[!htb]
\includegraphics[width=0.95\linewidth]{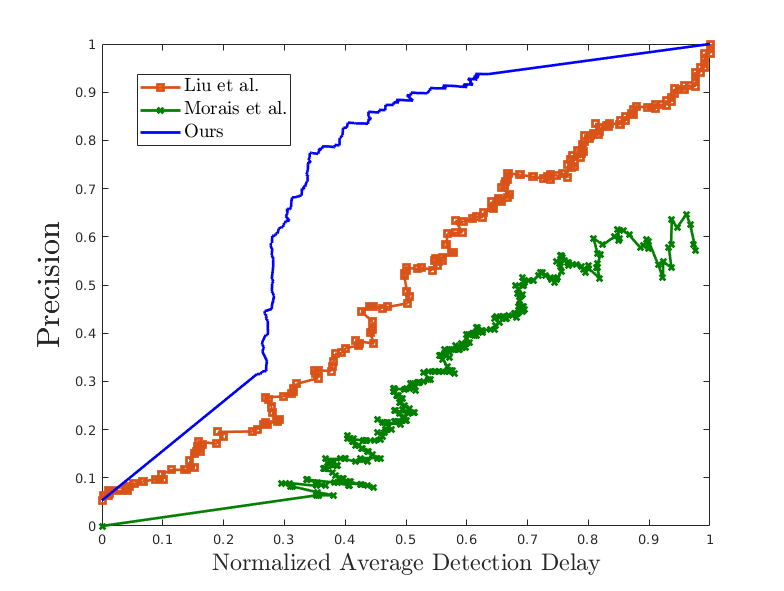}
\caption{Comparison of the proposed and the state-of-the-art algorithms Liu et al. \cite{liu2018future} and Morais et al. \cite{morais2019learning} in terms of online detection capability. The proposed algorithm has a significantly higher precision for any given detection delay.}
\label{f:prec_delay}
\vspace{-2mm}
\end{figure}

\textbf{Online Detection:} Since the proposed online detection formulation is event-based as compared to frame-based, it only considers an anomaly as a single event irrespective of the duration over which it occurs. In this setup, we present our results only on the ShanghaiTech dataset as the UCSD and CUHK Avenue datasets have fewer than 50 anomalous events, which is not enough for a reliable average performance comparison. A common technique used by several recent works \cite{liu2018future,ionescu2019object,morais2019learning,park2020learning} is to normalize the computed statistic for each test video independently, including the ShanghaiTech dataset. However, this methodology cannot be implemented in an online (real-time) system as it requires the prior knowledge of the minimum and maximum values the statistic might take. Moreover, many recent methods \cite{ionescu2019object,lu2020few,pang2020self} do not have their implementation details/code publicly available, while others are end-to-end \cite{pang2020self,ramachandra2020street,rodrigues2020multi} and cannot be implemented to work in an online fashion. Hence, we compare our method with the online versions of \cite{liu2018future,morais2019learning,luo2019video}. As shown in Fig. \ref{f:prec_delay}, our proposed algorithm achieves a better performance than the other algorithms in terms of quick detection and achieving high precision in alarms. This result is also summarized in Table \ref{tab:online} in terms of the APD values.    

\begin{table}[]
\centering
\begin{tabular}{c|c}
\hline
\multicolumn{2}{c}{\textbf{Online Detection}} \\ \hline \hline
\textbf{Methodology}   & \textbf{APD} \\ \hline \hline
Liu et al. \cite{liu2018future}    & 0.504 \\ \hline
Morais et al. \cite{morais2019learning}   & 0.324 \\ \hline
Luo et al. \cite{luo2019video} & 0.447 \\ \hline
\textbf{Ours} & \textbf{0.705} \\ \hline
\end{tabular}
\vspace{2mm}
\caption{Online detection comparison in terms of the proposed APD metric on the ShanghaiTech dataset. Higher APD value represents a better online anomaly detection performance.}
\label{tab:online}
\vspace{-3mm}
\end{table}

\begin{figure}[!htb]
\includegraphics[width=0.95\linewidth]{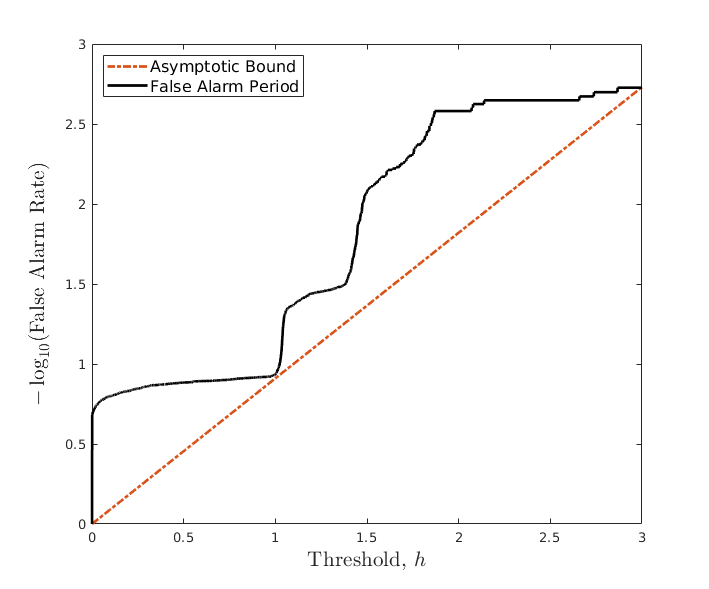}
\caption{Threshold selected according to Eq. \eqref{e:thr} satisfies the desired lower bound on false alarm period (i.e., upper bound on false alarm rate) even in the non-asymptotic regime with the finite sample size of the CUHK Avenue dataset.}
\label{f:thr}
\vspace{-2mm}
\end{figure}

\textbf{Threshold Selection:} We next evaluate the non-asymptotic use of the asymptotic threshold expression given in Eq. \eqref{e:thr}. As shown in Fig. \ref{f:thr}, even with the limited data size of the CUHK Avenue dataset, the derived expression satisfies the desired upper bound on the false alarm rate, which corresponds to a lower bound on the false period (inverse rate) in the figure.

\begin{table}[]
\centering
\resizebox{0.5\textwidth}{!}{%
\begin{tabular}{c|c|c|c}
\hline
\multicolumn{4}{c}{\textbf{Offline Localization}} \\ \hline \hline
\textbf{Methodology}    & \textbf{CUHK Avenue} &\textbf{ UCSD Ped 2 }& \textbf{ShanghaiTech}  \\ \hline \hline
MPPCA \cite{kim2009observe} & -           & 69.3       & -           \\ \hline
Del et al. \cite{del2016discriminative}     & 78.3        & -          & -   \\ \hline
Conv-AE \cite{hasan2016learning}        & 80.0        & 85.0       & 60.9\\ \hline
ConvLSTM-AE\cite{luo2017remembering}    & 77.0        & 88.1       & -   \\ \hline
Growing Neural Gas \cite{sun2017online}    & -           & 93.5       & -   \\ \hline
Stacked RNN\cite{luo2017revisit}    & 81.7        & 92.2       & 68.0\\ \hline
Deep Generic \cite{hinami2017joint}   & -           & 92.2       & -   \\ \hline
GANs \cite{ravanbakhsh2018plug}           & -           & 88.4       & -   \\ \hline
Future Frame \cite{liu2018future}     & 85.1        & 95.4       & 72.8\\ \hline
Skeletal Trajectory \cite{morais2019learning} & - & - & 73.4 \\ \hline
Multi-timescale Prediction \cite{rodrigues2020multi} & 82.85 & - & \textbf{76.03} \\ \hline
Memory-guided Normality \cite{park2020learning} & 88.5 & 97.0 & 70.5 \\ \hline

\textbf{Ours}          & \textbf{88.7}        & \textbf{97.2}       & 73.62     \\
\hline
\end{tabular}%
}
\caption{Offline anomaly localization comparison in terms of frame-level AUC on three datasets.}
\label{tab:offline}
\vspace{-4mm}
\end{table}





\begin{table*}[]
  \centering

  \resizebox{\textwidth}{!}{
  \begin{tabular*}{\textwidth}{c @{\extracolsep{\fill}}cccccccc}
    \hline
\textbf{Target} & \textbf{Methods} & \textbf{1-shot (K=1)} & \textbf{5-shot (K=5)} & \textbf{10-shot (K=10)}\\ 
\hline\hline
UCSD Ped 2& Pre-trained (ShanghaiTech) & 81.95 & 81.95 & 81.95\\
          & Pre-trained (UCF Crime) & 62.53 & 62.53 & 62.53\\
          & r-GAN (ShanghaiTech) & 91.19 & 91.8 & 92.8\\
          & r-GAN (UCF Crime) & 83.08 & 86.41 & 90.21\\
          & \textbf{Ours} & \textbf{93.19} & \textbf{95.91} & \textbf{96.01} \\
\hline
CUHK Avenue & Pre-trained (ShanghaiTech) & 71.43 & 71.43 & 71.43\\
            & Pre-trained (UCF Crime) & 71.43 & 71.43 & 71.43\\
            & r-GAN (ShanghaiTech) & 76.58 & 77.1 & 78.79\\ 
            & r-GAN (UCF Crime) & 72.62 & 74.68 & 79.02\\ 
            & \textbf{Ours} & \textbf{80.18}  & \textbf{80.21}   & \textbf{80.68} \\
\hline
UR Fall & Pre-trained (ShanghaiTech) & 64.08 & 64.08 & 64.08 \\ 
        & Pre-trained (UCF Crime) & 50.87 & 50.87 & 50.87 \\ 
        & r-GAN (ShanghaiTech) & 75.51 & 78.7 & 83.24\\
        & r-GAN (UCF Crime) & 74.59 & 79.08 & 81.85\\
        & \textbf{Ours} & \textbf{86.11} & \textbf{88.7} & \textbf{91.28}\\
 \hline
\end{tabular*}
}
\caption{Few-shot scene adaptation comparison of the proposed and the state-of-the-art \cite{lu2020few} algorithms in terms of frame-level AUC. The proposed algorithm is able to quickly adapt to new scenarios.}
\label{tab:fewshot}
\vspace{-3mm}
\end{table*}

\textbf{Offline Localization:} To show the offline localization capability of our algorithm, we also compare our algorithm to a wide range of state-of-the-art methods, as shown in Table \ref{tab:offline}, using the frame-level AUC criterion. The pixel-level criterion, which focuses on the spatial localization of anomalies, can be made equivalent to the frame-level criterion through simple post-processing techniques \cite{ramachandra2020street}. Hence, for offline anomaly localization, we consider frame-level AUC criterion. 
While \cite{ionescu2019object} recently showed significant gains over the other algorithms, their methodology of computing the average AUC over an entire dataset gave them an unfair advantage. Specifically, as opposed to determining the AUC on the concatenated videos, first the AUC for each video segment was computed and then those AUC values were averaged. 
As shown in Table \ref{tab:offline}, our proposed algorithm outperforms the existing algorithms on the UCSD Ped 2 and CUHK Avenue datasets, and performs competitively on the ShanghaiTech dataset. The multi-timescale framework \cite{rodrigues2020multi} is the only one that outperforms ours on the ShanghaiTech dataset since the anomalies are mostly caused by previously unseen human poses and \cite{rodrigues2020multi} extensively monitors them using a past-future trajectory prediction based framework. However, this causes their performance to severely degrade on the CUHK Avenue dataset, and similar to \cite{morais2019learning}, they cannot work on the UCSD dataset.

\textbf{Few-Shot Scene Adaptation:}
Our goal here is to compare the few-shot scene adaptation capability of the proposed algorithm and see how well it can generalize to new scenarios. In this case, we only use a few scenes from a specific scenario to adapt. However, few-shot scene adaptation is mostly unexplored and to the best of our knowledge only \cite{lu2020few} discusses it. Hence, following the experimental setup defined in \cite{lu2020few}, we use K-shots to adapt to a new scenario, where 1-shot is a sequence of 10 frames. From \cite{lu2020few}, we use the following baselines for comparison.

\textit{Pre-trained}: This baseline learns the model from videos available during training, then directly applies the model in testing without any adaptation.

\textit{r-GAN}: We also compare with a few-shot scene-adaptive anomaly detection model using a meta-learning framework proposed in \cite{lu2020few}. They use a GAN-based framework similar to \cite{liu2018future} and MAML algorithm for meta-learning. 

As compared to the pre-trained and r-GAN models, which need considerable training on either the ShanghaiTech or UCF Crime \cite{sultani2018real} dataset, our transfer learning based algorithm (pre-trained on generic datasets such as MS-COCO) is able to leverage our optical flow model which requires minimal computation to establish a baseline and adapt the decision parameter $h$ to a new scene. Due to the lack of available training data, we are unable to use the local motion and appearance features meaningfully, and hence our features are only dependant on the optical flow statistics. However, as shown in Table \ref{tab:fewshot}, we are still able to outperform the compared methods in terms of the frame-level AUC.

\begin{figure}[tbh]
\centering
  \subfloat[][\centering CUHK Avenue]{\includegraphics[width=0.98\linewidth]{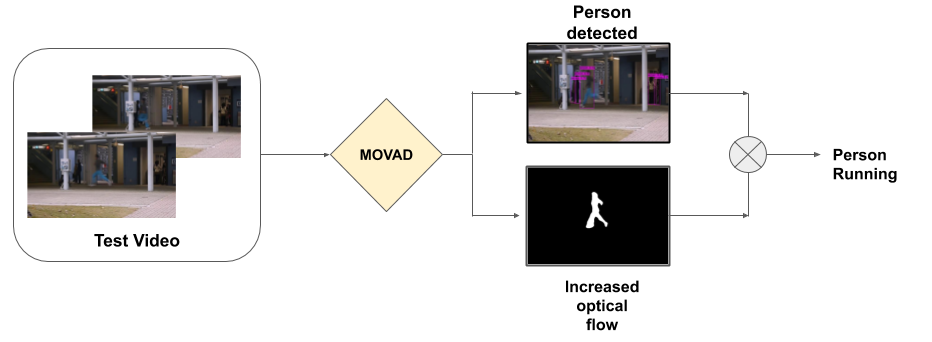}}
  \qquad
  \subfloat[][\centering UCSD]{\includegraphics[width=1.01\linewidth]{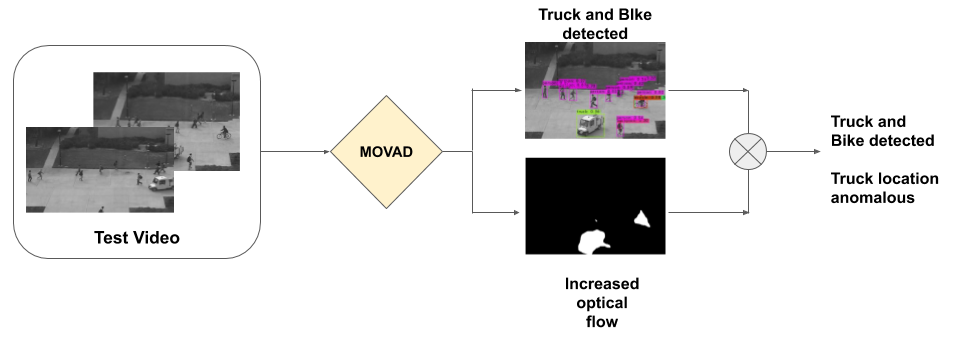}}
\caption{The proposed model is able to interpret the cause of the anomaly correctly.}
\label{f:sample}
\end{figure}

\begin{figure}
\vspace{-4mm}
\includegraphics[width=\linewidth]{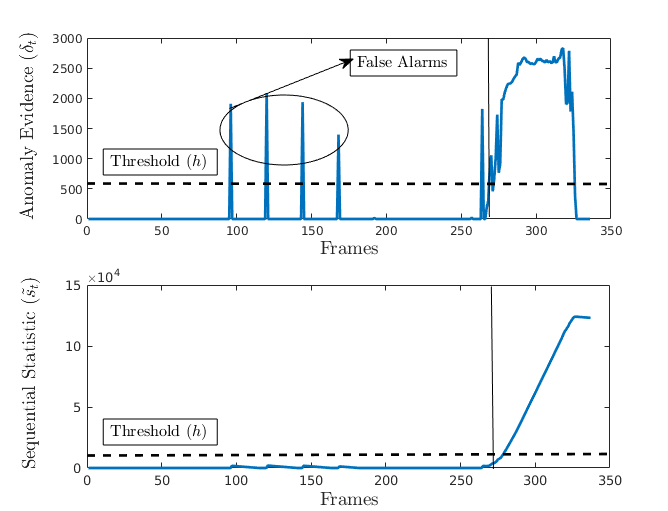}
\caption{The advantage of sequential anomaly detection over a single-shot detector. It is seen that a sequential detector can significantly reduce the number of false alarms.}
\label{f:sequen}
\vspace{-2mm}
\end{figure}
\vspace{-4mm}

\subsection{Ablation Study}

\begin{table}[]
\centering
\begin{tabular}{ccccc}
\hline
\multicolumn{2}{c}{\textbf{ShanghaiTech }} \\ \hline \hline
\textbf{Module}   & \textbf{AUC} \\ \hline \hline
Object Detection    & 0.594 \\ \hline
Optical Flow   & 0.703 \\ \hline
Pose Estimation & 0.652 \\ \hline
\end{tabular}
\caption{Performance of each module in terms of the frame-level AUC on the ShanghaiTech dataset.}
\label{tab:ablation}
\vspace{-4mm}
\end{table}

In Table \ref{tab:ablation}, we present the results for each module of the proposed MOVAD framework on the ShanghaiTech dataset. While it is clear that optical flow is the major contributor among all the modules in this dataset, each module serves a specific purpose. In this dataset, although several recent works perform closely to the proposed framework, a distinguishing advantage of MOVAD is its interpretability. By leveraging the statistical nature of our decision making module, it is possible to determine the cause of increase in the decision statistic. In Fig. \ref{f:sample}, we present a sample scenario from the CUHK Avenue and UCSD datasets, in which the proposed detector is able to evaluate the statistics from each module and justify the cause of the anomaly. However, since there is no ground truth available in terms of the description of the anomaly, we were unable to quantitatively evaluate the interpretability performance of MOVAD.

\textbf{Impact of Sequential Detection:} To emphasize the significance of the proposed sequential detection method, we compare a nonsequential version of our algorithm by applying a threshold to the instantaneous anomaly evidence $\delta_t$ (Sec. \ref{s:sad}), which is similar to the approach employed by many recent works \cite{liu2018future,sultani2018real,ionescu2019object}. As shown in Fig. \ref{f:sequen}, the proposed sequential statistic handles noisy evidence by integrating recent evidence over time. On the other hand, the instantaneous anomaly evidence is more prone to false alarms since it only considers the noisy evidence available at the current time to decide. Specifically, without sequential detection, the APD presented in Table \ref{tab:online} for the proposed framework reduces to 0.673.

\vspace{-2mm}
\section{Conclusion and Discussions}
\vspace{-2mm}
For video anomaly detection, we presented a modular framework called MOVAD, which consists of an interpretable transfer learning based feature extractor, and a novel $k$NN-RNN based sequential anomaly detector. Mathematical analysis was provided for false alarm rate and threshold selection. 
Following the timely detection requirement in practical settings, MOVAD first detects anomalous events in an online fashion, and then deals with localizing the anomalous video frames. Online detection of anomalous events is largely overlooked in the video anomaly detection literature, thus a new performance metric was also introduced to compare algorithms in terms of online anomaly detection in videos. Through extensive testing on the benchmark datasets, we show that MOVAD significantly outperforms the state-of-the-art methods for online detection while performing competitively for offline localization. 

While being able to capture anomalies in various video aspects, such as object appearance and motion, the proposed method currently is not optimized for specific anomaly types. For instance, it is not able to detect unexpected human poses as the optical flow does not change significantly (see Supplementary). For future work, we plan to focus on continual and self-supervised learning for MOVAD. 

{\small
\bibliographystyle{ieee_fullname}
\bibliography{egbib}
}

\end{document}